\documentclass[10pt,twocolumn,letterpaper]{article}

\usepackage{cvpr}
\usepackage{times}
\usepackage{epsfig}
\usepackage{graphicx}
\usepackage{amsmath}
\usepackage{amssymb}
\usepackage{stmaryrd}
\usepackage{multirow}
\usepackage{graphicx}
\usepackage{adjustbox}

\usepackage[pagebackref=true,breaklinks=true,letterpaper=true,colorlinks,bookmarks=false]{hyperref}

\cvprfinalcopy 

\def\cvprPaperID{****} 

\ifcvprfinal\fi
\begin{document}

\title{DepthNet Nano: A Highly Compact Self-Normalizing Neural Network for Monocular Depth Estimation}

\author{Linda Wang, Mahmoud Famouri, and Alexander Wong\\
Department of Systems Design Engineering, University of Waterloo, Canada\\
Waterloo Artificial Intelligence Institute, Canada\\
DarwinAI Corp., Canada\\
{\tt\small \{linda.wang, alexander.wong\}@uwaterloo.ca}
}

\maketitle

\begin{abstract}
\vspace{-10pt}
Depth estimation is an active area of research in the field of computer vision, and has garnered significant interest due to its rising demand in a large number of applications ranging from robotics and unmanned aerial vehicles to autonomous vehicles. A particularly challenging problem in this area is monocular depth estimation, where the goal is to infer depth from a single image.  An effective strategy that has shown considerable promise in recent years for tackling this problem is the utilization of deep convolutional neural networks.  Despite these successes, the memory and computational requirements of such networks have made widespread deployment in embedded scenarios very challenging. In this study, we introduce DepthNet Nano, a highly compact self normalizing network for monocular depth estimation designed using a human machine collaborative design strategy, where principled network design prototyping based on encoder-decoder design principles are coupled with machine-driven design exploration. The result is a compact deep neural network with highly customized macroarchitecture and microarchitecture designs, as well as self-normalizing characteristics, that are highly tailored for the task of embedded depth estimation. The proposed DepthNet Nano possesses a highly efficient network architecture (e.g., $24\times$ smaller and $42\times$ fewer MAC operations than Alhashim et al. on KITTI), while still achieving comparable performance with state-of-the-art networks on the NYU-Depth V2 and KITTI datasets. Furthermore, experiments on inference speed and energy efficiency on a Jetson AGX Xavier embedded module further illustrate the efficacy of DepthNet Nano at different resolutions and power budgets (e.g., $\sim$14 FPS and $>$0.46 images/sec/watt at $384\times1280$ at a 30W power budget on KITTI).
\end{abstract}

\vspace{-10pt}
\section{Introduction}
\begin{figure}
\centering
  \includegraphics[width=\linewidth]{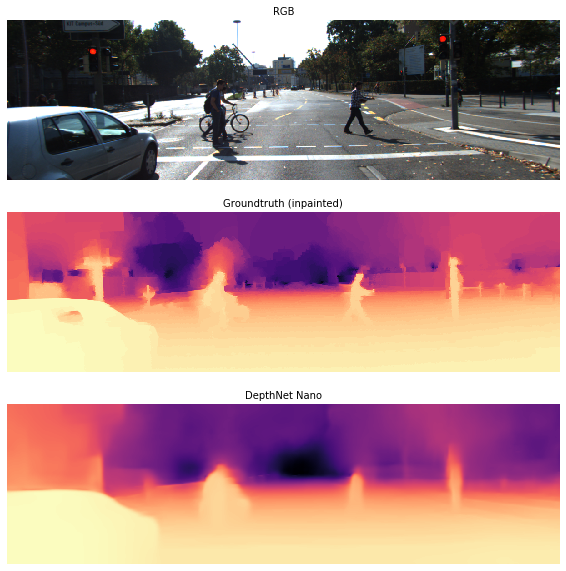}
  \caption{\textbf{Monocular depth estimation.} Top: RGB image (from KITTI dataset), Middle: Corresponding ground-truth dense depth map.
  Bottom: Estimated dense depth map by Depth-Net Nano (designed using the human-machine collaborative design strategy).}
  \label{fig:example-depth}
  \vspace{-10pt}
\end{figure}

The task of estimating depth from 2D images is crucial for applications such as 3D scene understanding and reconstruction. Recently, monocular depth estimation, where a dense depth map is obtained from a single image, as shown in Figure \ref{fig:example-depth}, has gained traction. Compared to depth estimation from stereo images or video sequence, monocular depth estimation is an ill-posed problem, which means there are more than one possible unique solution. To tackle this ill-posed problem, one method that has shown promise is deep convolutional neural networks (DCNNs).

These DCNNs learn deep features to estimate a depth for each pixel. Most of the recent developments have focused on an encoder-decoder architecture with very powerful deep neural network backbone macroarchitecture designs, such as VGG, ResNet and DenseNet~\cite{fu2018deep, alhashim2018high, laina2016deeper}, to learn deep features. However, these current deep neural networks for monocular depth estimation are large and difficult to deploy and run in edge scenarios such as robots, drones, and autonomous vehicles. In addition to possessing very high network architecture complexity, these current deep neural networks have high computation time and low energy efficiency, which makes them difficult to deploy for mission-critical scenarios such as self-driving cars, where reaction time is crucial for driver safety.

Taking inspiration from recent work in efficient object detection, where large backbone architectures are replaced with more efficient backbone network architectures, such as MobileNetV2~\cite{sandler2018mobilenetv2}, Wofk et al.~\cite{wofk2019fastdepth} used smaller backbone architectures (e.g., ResNet-16 and MobileNet) in order to decrease the number of parameters and run-time necessary to operate on embedded devices.  To further reduce the network size and inference runtime, network pruning was applied ~\cite{yang2018netadapt}. While the resulting depth estimation networks achieved significant improvements in terms of inference speed, the depth estimation performance was significantly lower and not comparable with current state of the art.

Taking a different direction than manual architecture selection strategies and network pruning strategies, human-machine collaborative design strategies~\cite{wong2018ferminets} has shown recent success in designing highly compact DCNNs by coupling principled network design prototyping and machine-driven design exploration based on human-specified design requirements and constraints. In particular, such strategies have been demonstrated to be quite effective at designing efficient deep neural networks well suited for various perception tasks, such as object detection, image classification, and semantic segmentation~\cite{wong2019yolo, wong2018ferminets}.

\begin{figure}
  \includegraphics[width=\linewidth]{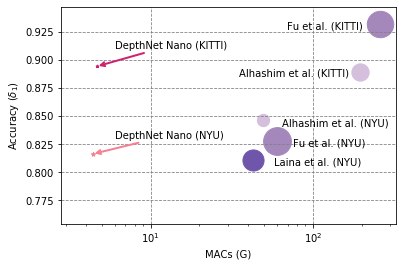}
  \caption{\textbf{Accuracy vs. MACs and number of parameters} comparison of various depth estimation networks. The size of the point represents the number of parameters in the depth estimation network. Top left represents high accuracy and high efficiency.}
  \label{fig:comp-plot}
\end{figure}

In this study, we explore a human-machine collaborative design strategy to design highly compact deep convolutional neural networks for the task of monocular depth estimation on the edge. More specifically, we leverage encoder-decoder design principles that were found to be effective in current state of the art monocular depth estimation to create DepthNet Nano, a highly compact network with highly customized module-level macroarchitecture and microarchitecture designs tailored specifically for embedded depth estimation. In Figure \ref{fig:comp-plot}, we show that DepthNet Nano is significantly smaller and faster than current state of the art depth estimation networks while maintaining a comparable accuracy across two widely-used benchmark datasets.

The paper is organized as follows.  Section 2 provides a detailed description of the human-machine collaborative design strategy leveraged in this study to create DepthNet Nano.  Section 3 provides a detailed description and discussion of interesting characteristics of the resulting architecture design of DepthNet Nano.  Section 4 presents and discusses the results from the quantitative and qualitative experiments conducted to study the efficacy of DepthNet Nano when compared to state-of-the-art depth estimation networks.  Finally, Section 5 draws conclusions and discusses future directions.

\section{Methods}
In this study, we leverage a human-machine collaborative design strategy to design DepthNet Nano, a highly compact DCNN tailored for monocular depth estimation under edge scenarios. The human-machine collaborative design strategy comprises of two main design stages: i) principled network design prototyping, and ii) machine-driven design exploration. 

\subsection{Principled Network Design Prototyping}
The network design prototyping stage is the initial design stage, where we create an initial network design prototype (denoted as $\varphi$) based on human-driven design principles to guide the machine-driven design exploration stage. In this study, an initial network design prototype was constructed based on densely-connected encoder-decoder architecture design principles~\cite{alhashim2018high}, which has been demonstrated to be quite successful in achieving high-resolution monocular depth estimation.  

A standout characteristic of the densely-connected encoder-decoder architecture is the leveraging of a large number of direct connections between not only encoder layers, but also between encoder and decoder layers.  Below is a detailed description of the design principles leveraged in constructing the initial network design prototype.  It is very important to note that the actual macroarchitecture and microarchitecture designs of the individual modules and layers in the final DepthNet Nano network architecture, as well as the number of network modules, are left for the machine-driven design exploration stage to decide in an automatic manner based on both human-specified design requirements and constraints catered to edge device scenarios with limited computational and memory capabilities. 

\subsubsection{Encoder-Decoder Architecture}
In this study, the initial network design prototypes possess an encoder-decoder architecture that is designed for dense depth map generation.  The encoder layers in such an architecture are designed to learn a multitude of low to mid level features for characterizing an input scene.  Next, the decoder layers are designed to merge and upsample the features learned from the encoder layers to recover a dense depth map. The decoder layers consist of upsampling blocks  followed by a concatenation and two convolutional operations. Finally, the encoder-decoder architecture leveraged by the initial network design prototypes follows the decoder layers with a $3\times3$ convolutional layer at the end, which is designed to produce the final dense depth map. 

\subsubsection{Encoder-Encoder and Encoder-Decoder Skip Connections}
In general, as the number of layers increases, the network accuracy improves because each layer is learning deeper features. However, He et al.~\cite{he2016deep} found that a 56-layer deep convolutional neural network has higher training and test error than a 26-layer CNN deep convolutional neural network introduced a fundamental building block, the residual block, to alleviate training of deep neural networks. The residual block adds a previous layer to the current layer. By adding information from previous layers, the network can learn residuals or errors between a previous layer and the current one. Extending upon this idea of skip connections, densely-connected network architectures consists of a large number of skip connections between different layers.  More specifically, instead of adding the previous layer as an identity function, densely-connected architectures concatenate outputs from previous layers to the current layer. This is found to alleviate the vanishing gradient problem, strengthen feature propagation and feature reuse~\cite{huang2017densely}.  A such, the introduction of encoder-encoder skip connections into a deep encoder-decoder architecture can improve the training process and improve network performance.

Furthermore, in the case of encoder-decoder architectures, as the layers in the encoder get deeper, higher level features are learned; however, the resolution of the feature maps get progressively lower. As such, the input to the decoder is of low resolution.  Since the purpose of the decoder network is to upsample the features learned from the encoder, the resulting depth map image would also have of low resolution. 

To overcome the low resolution problem, an effective strategy is the leveraging of skip connections between encoder and decoder layers within an encoder-decoder architecture. Such encoder-decoder skip connections merge high resolution feature maps from the encoder layers to the features in the decoder layers, resulting in a more detailed decoder output. 

\subsection{Machine-driven Design Exploration}
The machine-driven design exploration stage takes in the given data, initial network design prototype $\varphi$, and human-specified design requirements and constraints, which are designed specifically around edge scenarios with limited computational and memory capabilities.

Using the initial network design prototype $\varphi$ described in the previous section, as well as human specified design requirements, a machine-driven design exploration is leveraged in the form of generative synthesis~\cite{wong2018ferminets} to determine macroarchitecture and microarchitecture designs for depth estimation on edge devices. The process of generative synthesis is capable of determining the optimal network macroarchitecture and microarchitecture design that satisfy the human-specified constraints. This is achieved by learning generative machines that can generate deep neural networks that meet the specified constraints. To learn the optimal generator, generative synthesis is formulated as a constrained optimization problem, defined in Equation \ref{eq:gensynth}, where given a set of seeds $\mathcal{S}$, a generator $\mathcal{G}$ can generate networks $\{\mathcal{N}_s | s \in \mathcal{S}\}$ that maximize a universal performance function $\mathcal{U}$, while also satisfying constraints defined in an indicator function $1_r(\cdot)$.
\begin{equation}
    \mathcal{G} = \max_{\mathcal{G}} \mathcal{U}(\mathcal{G}(s)) \text{ subject to } 1_r(\mathcal{G}(s)) = 1, \forall s \in \mathcal{S}
    \label{eq:gensynth}
\end{equation}

The generative synthesis process is guided by both the initial prototype $\varphi$ and human-specified constraints. To guide the process towards learning generative machines that generate highly efficient and compact depth estimation networks for edge devices, an indicator function $1_r(\cdot)$ is configured so that the generated networks are within the human-specified constraints. In this study, for generating a highly efficient and compact depth estimation network tailored for NYU Depth v2, the indicator function  $1_r(\cdot)$ was set up such that: i) $\delta_1$ accuracy $\geq$ 0.81 on NYU Depth v2, and ii) network architecture complexity $\leq$ 5M parameters.  The $\delta_1$ accuracy and network architecture complexity conditions in the indicator function $1_r(\cdot)$ are set for this case such that the $\delta_1$ accuracy of the resulting DepthNet Nano network exceeds that of Laina et al.~\cite{laina2016deeper}, a popular  deep convolutional neural network for monocular depth estimation for NYU Depth v2, while having more than 8$\times$ fewer parameters.  

For generating a highly efficient and compact depth estimation network tailored for KITTI, the indicator function  $1_r(\cdot)$ was set up for this case such that: i) $\delta_1$ accuracy $\geq$ 0.89 on KITTI, and ii) architectural complexity $\leq$ 2M parameters.  The $\delta_1$ accuracy and network architecture complexity conditions in the indicator function $1_r(\cdot)$ are set such that the $\delta_1$ accuracy of the resulting DepthNet Nano network exceeds to that of Alhashim et al.~\cite{alhashim2018high}, a popular, high-performance deep convolutional neural network for monocular depth estimation for KITTI, while having more than 20$\times$ fewer parameters.

Finally, in this study, the universal performance function $\mathcal{U}$ leveraged in Eq.~\ref{eq:gensynth} is NetScore~\cite{wong2019netscore}, a quantitative performance metric designed for assessing the balance between accuracy, computational complexity, network architecture complexity of a deep neural network. The NetScore metric is defined as
\begin{equation}
\label{netscore}
\Omega(\mathcal{N}) = 20\log\Big(\frac{a(\mathcal{N})^\kappa}{p(\mathcal{N})^\beta r(\mathcal{N})^\gamma}\Big)
\end{equation} 
\noindent where for this study, $a(\mathcal{N})$ is the combination of $\delta_1$ accuracy and absolute relative error, $p(\mathcal{N})$ is the number of parameters in the network, $m(\mathcal{N})$ is the number of multiply-accumulate (MAC) operations, and $\kappa$, $\beta$, $\gamma$ control the the influence of accuracy, architectural complexity and computational complexity, respectively. For this study, $\kappa$ is set to $0.7$, $\beta$ and $\gamma$ are both set to $0.15$ to put an emphasis on accuracy while maintaining balance with architectural complexity and computational complexity.

\section{DepthNet Nano Architectural Design}
\begin{figure*}
  \includegraphics[width=\textwidth]{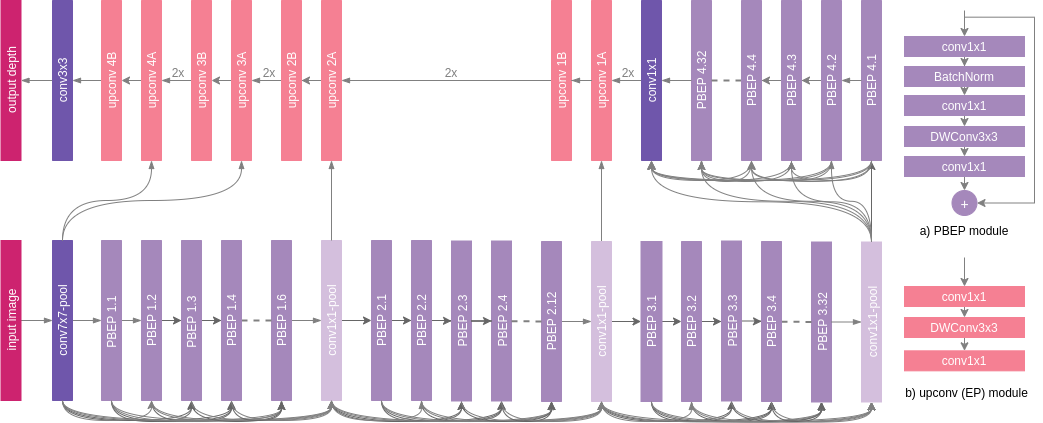}
  \caption{\textbf{DepthNet Nano Architecture.} The network architecture exhibits high macroarchitecture heterogeneity with a mix of PBEP and EP modules, as well as individual 7$\times$7, 3$\times$3, and pointwise convolution layers.  Furthermore, the architecture also exhibits high microarchitecture heterogeneity, as seen in Table~\ref{table:arch}. Finally, the network architecture possesses a very deep densely-connected self-normalization macroarchitecture, which has not been previously explored.}
  \label{fig:arch}
\end{figure*}

The network architecture of the proposed DepthNet Nano, which is illustrated in Fig.~\ref{fig:arch} with microarchitecture details provided in Table~\ref{table:arch}, has several interesting characteristics that are discussed in detail below.

\subsection{Self-Normalization Macroarchitecture}
The first interesting characteristic of the DepthNet Nano architecture is its self-normalizing property within a very deep densely-connected network architecture, which has not been previously explored. More specifically, rather than leveraging popular activation functions such as Rectifier Linear Units (ReLU) that are more commonly found in depth estimation networks, the proposed DepthNet Nano architecture heavily leverages Scaled Exponential Linear Units (SELU)~\cite{klambauer2017self} as the only form of activation within the network architecture, which can be defined as

\begin{equation}
\text{selu}(x) = \lambda
  \begin{cases}
    x & \text{if } x>0 \\
    \alpha\text{exp}(x)-\alpha & \text{if } x\leq0
  \end{cases}
\label{eq:selu}
\end{equation}

One of the key advantages of SELUs is their self-normalizing properties that makes learning more robust for deep neural networks.  More specifically, since the activations in each layer of the network are close to zero mean and unit variance, as the data is propagated through, it will converge towards zero mean and unit variance. The self-normalizing property is achieved with SELUs by decreasing the variance for negative inputs and increasing the variance for positive inputs. To achieve zero mean and unit variance, the amount of decrease for very negative inputs and the amount of increase for near zero values are greater than other inputs.  The result is a self-normalizing neural network that is able to achieve high depth estimation accuracy while being extremely efficient and possessing very low network architecture complexity despite high network inter-connectivity.

\subsection{Densely Connected Projection BatchNorm Expansion Projection Macroarchitecture}
Another interesting characteristic of the DepthNet Nano architecture is the densely connected projection batchnorm expansion projection (PBEP) module, which are leveraged heavily in the encoding layers of the network architecture (see Fig. \ref{fig:arch}). Compared to the expansion projection (EP) modules leveraged extensively in the decoding layers of the network architecture (see Fig. \ref{fig:arch}), which have been seen in other literature on efficient network architectures~\cite{sandler2018mobilenetv2,mnas,fairnas}, PBEP has an additional projection layer that decreases the number of channels of the previous layer before expanding the layer for depth-wise convolution. PBEP macroarchitecture consists of:
\begin{enumerate}
  \item A projection layer, where the output channels of the previous layer are projected to a lower dimensionality in this layer using $1\times1$ convolutions.
  \item A batch normalization layer that normalizes the output of a previous layer to improve the stability of the network.
  \item An expansion layer, where $1\times1$ convolutions are leveraged to expand the output of the batch normalization layer to a higher dimensionality.
  \item A depth-wise convolution layer, where spatial convolutions with a different filter are applied to each of the individual output channels of the expansion layer.
  \item A projection layer with $1\times1$ convolutions that projects the output channels from the depth-wise convolution layer to a lower dimensionality.
\end{enumerate}
The use of densely connected PBEP macroarchitectures reduces the architectural and computational complexity of the DepthNet Nano architecture while maintaining high model expressiveness and producing high quality depth estimations.

\subsection{Macroarchitecture and Microarchitecture Heterogeneity}
Unlike hand-crafted architectures, the generated macroarchitecture and microarchitecture within the network can differ greatly from layer to layer. There are a mix of different type of modules, such as PBEP and EP modules, as well as individual $7\times7$, $3\times3$ and $1\times1$ convolution layers. In addition, the same module type has vastly different microarchitectures since each module is catered specifically for the needs of the task. For instance, each PBEP and EP module have different numbers of channels to represent the learned features and a different multiplicity for the channel expansion layer.

The benefit of high macroarchitecture and microarchitecture heterogeneity in DepthNet Nano network architecture is that it enables each component of the network architecture to be uniquely tailored to achieve a very strong balance between architectural and computation complexity while maintaining model expressiveness. The architectural diversity in DepthNet Nano demonstrates the advantage of leveraging a human-collaborative design strategy as it would be difficult for a human designer, or other design exploration methods to customize a network architecture to the same level of architectural granularity.

\begin{table}[]
\centering
\caption{\textbf{DepthNet Nano microarchitecture details for NYU Depth v2 and KITTI.}}
\begin{adjustbox}{max width=\linewidth}
\begin{tabular}{c|c|c}
\hline
\textbf{Layer} & \textbf{NYU} & \textbf{KITTI} \\ \hline \hline
input & $480\times640\times3$ & $384\times1280\times3 $ \\ \hline
conv$7\times7$-pool & $240\times320\times15$ & $192\times640\times14$ \\ \hline
PBEP 1.1 & $120\times160\times16$ & $96\times320\times13$ \\ \hline
PBEP 1.2 & $120\times160\times16$ & $96\times320\times15$ \\ \hline
PBEP 1.3 & $120\times160\times18$ & $96\times320\times18$ \\ \hline
PBEP 1.4 & $120\times160\times20$ & $96\times320\times13$ \\ \hline
...            & ...          & ...           \\ \hline
PBEP 1.6 & $120\times160\times20$ & $96\times320\times17$ \\ \hline
conv$1\times1$-pool & $60\times80\times57$ & $48\times160\times30$ \\ \hline
PBEP 2.1 & $60\times80\times11$ & $48\times160\times9$ \\ \hline
PBEP 2.2 & $60\times80\times15$ & $48\times160\times13$ \\ \hline
PBEP 2.3 & $60\times80\times19$ & $48\times160\times13$  \\ \hline
PBEP 2.4 & $60\times80\times19$ & $48\times160\times18$  \\ \hline
...            & ...          & ...           \\ \hline
PBEP 2.12 & $60\times80\times22$ & $48\times160\times19$ \\ \hline
conv$1\times1$-pool & $30\times40\times133$ & $24\times80\times79$ \\ \hline
PBEP 3.1 & $30\times40\times17$ & $24\times80\times13$ \\ \hline
PBEP 3.2 & $30\times40\times16$ & $24\times80\times14$ \\ \hline
PBEP 3.3 & $30\times40\times16$ & $24\times80\times18$ \\ \hline
PBEP 3.4 & $30\times40\times17$ & $24\times80\times15$ \\ \hline
...            & ...          & ...           \\ \hline
PBEP 3.32 & $30\times40\times20$ & $24\times80\times14$ \\ \hline
conv$1\times1$-pool & $15\times20\times336$ & $12\times40\times117$ \\ \hline
PBEP 4.1 & $15\times20\times19$ & $12\times40\times17$ \\ \hline
PBEP 4.2 & $15\times20\times15$ & $12\times40\times13$ \\ \hline
PBEP 4.3 & $15\times20\times19$ & $12\times40\times14$ \\ \hline
PBEP 4.4 & $15\times20\times18$ & $12\times40\times12$ \\ \hline
...            & ...          & ...           \\ \hline
PBEP 4.32 & $15\times20\times15$ & $12\times40\times11$ \\ \hline
conv$1\times1$ & $15\times20\times302$ & $12\times40\times176$         \\ \hline
upconv 1A & $30\times40\times138$ & $24\times80\times86$ \\ \hline
upconv 1B & $30\times40\times118$ & $24\times80\times112$ \\ \hline
upconv 2A & $60\times80\times54$ & $48\times160\times47$ \\ \hline
upconv 2B & $60\times80\times40$ & $48\times160\times48$ \\ \hline
upconv 3A & $120\times160\times23$ & $96\times320\times28$ \\ \hline
upconv 3B & $120\times160\times16$ & $96\times320\times25$ \\ \hline
upconv 4A & $240\times320\times11$ & $192\times640\times17$ \\ \hline
upconv 4B & $240\times320\times12$ & $192\times640\times24$ \\ \hline
conv$3\times3$ & $240\times320\times1$ & $192\times640\times1$ \\ \hline
output & $480\times640\times1$ & $384\times1280\times1$ \\ \hline
\end{tabular}
\label{table:arch}
\end{adjustbox}
\vspace{-10pt}
\end{table}

\section{Experimentation Results}
\begin{table*}
\centering
\caption{\textbf{Performance on KITTI.} All networks are evaluated using a pre-defined center cropping~\cite{eigen2014depth}. Best results in \textbf{bold}.}
\begin{adjustbox}{max width=\textwidth}
\begin{tabular}{c||c|c|c|c|c|c|c|c|c|c}
\hline
\multirow{2}{*}{Model} &
\multirow{2}{*}{Input Size} & \multirow{2}{*}{MACs [G]} & \multirow{2}{*}{Params [M]} & \multicolumn{3}{c|}{higher is better} & \multicolumn{4}{c}{lower is better} \\ \cline{5-11}
 &  &  &  & $\delta_1 < 1.25$ & $\delta_2 < 1.25^2$ & $\delta_3 < 1.25^3$ & Abs Rel & Sq Rel & RMSE & RMSE \textit{log} \\ \hline \hline
Fu et al.~\cite{fu2018deep} & $385\times513$ & 258 & 99.8 & \textbf{0.932} & \textbf{0.984} & \textbf{0.994} & \textbf{0.072} & \textbf{0.307} & \textbf{2.727} & \textbf{0.120} \\ \hline
Alhashim et al.~\cite{alhashim2018high} & $384\times1280$ & 196 & 42.8 & 0.886 & 0.965 & 0.986 & 0.093 & 0.589 & 4.170 & 0.171 \\ \hline \hline
DepthNet Nano & $384\times1280$ & \textbf{4.66} & \textbf{1.75} & 0.894 & 0.978 & \textbf{0.994} & 0.103 & 0.511 & 3.916 & 0.150 \\ \hline
\end{tabular}%
\label{tab:comp-perf-kitti}
\end{adjustbox}
\end{table*}

\begin{table*}
\centering
\caption{\textbf{Performance on NYU Depth v2.} All networks are evaluated using a pre-defined center cropping~\cite{eigen2014depth}. Best results in \textbf{bold}.}
\begin{adjustbox}{max width=\textwidth}
\begin{tabular}{c||c|c|c|c|c|c|c|c|c}
\hline
\multirow{2}{*}{Model} &
\multirow{2}{*}{Input Size} & \multirow{2}{*}{MACs [G]} & \multirow{2}{*}{Params [M]} & \multicolumn{3}{c|}{higher is better} & \multicolumn{3}{c}{lower is better} \\ \cline{5-10}
 &  &  &  & $\delta_1 < 1.25$ & $\delta_2 < 1.25^2$ & $\delta_3 < 1.25^3$ & Abs Rel & RMSE & Log10 \\ \hline \hline
Laina et al.~\cite{laina2016deeper} & $228\times304$ & 42.7 & 63.6 & 0.811 & 0.953 & 0.988 & 0.127 & 0.573 & 0.055 \\ \hline
Fu et al.~\cite{fu2018deep} & $257\times353$ & 60.2 & 110.3 & 0.828 & 0.965 & 0.992 & \textbf{0.115} & 0.509 & \textbf{0.051} \\ \hline
Alhashim et al.~\cite{alhashim2018high} & $480\times640$ & 49.2 & 21.52 & \textbf{0.846} & \textbf{0.974} & \textbf{0.994} & 0.123 & \textbf{0.465} & 0.053 \\ \hline \hline
DepthNet Nano & $480\times640$ & \textbf{4.4} & \textbf{3.46} & 0.816 & 0.958 & 0.989 & 0.139 & 0.599 & 0.059 \\ \hline
\end{tabular}%
\label{tab:comp-perf}
\end{adjustbox}
\end{table*}

To demonstrate the efficacy of the proposed DepthNet Nano network designed using the human-machine collaborative design strategy, we examine its network architecture complexity, depth estimation performance, and computational cost across two popular benchmark datasets.

\subsection{Implementation Details}
The proposed DepthNet Nano was implemented using the TensorFlow open source platform for machine learning.  The training scheme leveraged in this study was that outlined in \cite{alhashim2018high}. The ADAM optimizer~\cite{Adam} was leveraged with a learning rate of $0.00005$ and parameter values $\beta_1=0.9$ and $\beta_2=0.999$.

\subsection{Datasets}
Two benchmark datasets were evaluated to investigate the efficacy of the proposed DepthNet Nano: i) NYU-Depth V2, and ii) KITTI.  The NYU-Depth V2 dataset provides 2D images and corresponding depth maps captured using Microsoft Kinect of 464 unique indoor scenes~\cite{silberman2012indoor}. For this study, the networks are trained on 50,000 images, which is a subset of the NYU-Depth V2 dataset, and tested on 654 images, following the procedure established in \cite{alhashim2018high}. The KITTI dataset contains 2D outdoor scenes and corresponding ground-truth lidar points captured using a Velodyne HDL-64E~\cite{geiger2012we}. Since this study requires dense input depth maps, 23158 lidar scans were inpainted using Levin et al's colorization method~\cite{levin2004colorization}, as consistent with other studies. For this study, using the same procedure as previous literature, the networks are trained on 23158 color images and their corresponding dense depth maps, and tested on 697 images from 29 scenes split by Eigen et al.~\cite{eigen2014depth}. For testing, the depth predictions for are evaluated on a pre-defined center cropping by Eigen~\cite{eigen2014depth}.

\subsection{Performance Evaluation Metrics}
\label{sec:eval-metrics}
Each tested network in this study is evaluated using several error and accuracy metrics that were used in prior works~\cite{alhashim2018high, eigen2014depth, fu2018deep, laina2016deeper}. More specifically, the following performance metrics were leveraged in this study:
\begin{itemize}
  \item relative absolute error (Abs Rel): $\frac{1}{N} \sum_{i=1}^N \frac{|y_i - \hat{y}_i|}{\hat{y}_i}$
  \item relative squared error (Sq Rel):
  $\frac{1}{N} \sum_{i=1}^N \frac{(y_i - \hat{y}_i)^2}{\hat{y}_i}$
  \item root mean squared error (rmse): $\sqrt{\frac{1}{N} \sum_{i=1}^N (y_i - \hat{y}_i)^2}$
  \item log rmse (rmse \textit{log}):
  $\sqrt{\frac{1}{N} \sum_{i=1}^N (log(y_i) - log(\hat{y}_i))^2}$
  \item average log error (Log10): $\frac{1}{N} \sum_{i=1}^N |log(y_i) - log(\hat{y}_i)|$
  \item  $\delta_i$ accuracy: \% of $y_i$ s.t. $max(\frac{y_i}{\hat{y}_i}, \frac{\hat{y}_i}{y_i}) = \delta < thr$ for $thr = 1.25, 1.25^2, 1.25^3$
\end{itemize}
\noindent where $y_i$ is a pixel in predicted depth image, $\hat{y}_i$ is a pixel in the ground-truth depth image and $N$ is the total number of pixels in the depth image.

Finally, evaluating the real-world performance of the proposed DepthNet Nano in a realistic embedded scenario, we performed inference speed and power efficiency evaluations on a Jetson AGX embedded module at two different power budgets: 1) 30W and 2) 15W.  For inference speed, we computed the framerate (FPS), while for power efficiency we computed the number of images processes per sec per watt (i.e., images/sec/watt).

\subsection{Quantitative Analysis}
To study the efficacy of human-machine collaborative design, we evaluate DepthNet Nano alongside state-of-the-art depth estimation networks on performance metrics defined in Section \ref{sec:eval-metrics}, network architecture complexity, and computational complexity on the NYU Depth V2 and KITTI datasets, as shown in Table~\ref{tab:comp-perf} and Table~\ref{tab:comp-perf-kitti}, respectively. Since this study targets high quality depth estimation, only state-of-the-art networks in literature with $\delta_1$ accuracies greater than $0.8$ and $0.88$ are considered on NYU Depth v2 and KITTI, respectively.

\textbf{KITTI}. It was observed that DepthNet Nano had significantly lower architecture complexity and computational complexity compared to the tested state-of-the-art networks.  For example, DepthNet Nano has $>$24$\times$ fewer parameters and requires $>$42$\times$ fewer multiply-accumulate operations (MACs) for inference than~\cite{alhashim2018high}, while achieving higher $\delta_1$, $\delta_2$ and $\delta_3$ accuracies.

\textbf{NYU Depth v2}. It was observed that, as in the case of KITTI, DepthNet Nano had significantly lower architecture complexity and computational complexity compared to the tested state-of-the-art networks.  For example, DepthNet Nano has $>$18$\times$ fewer parameters and requires $>$9.7$\times$ fewer MAC operations for inference than~\cite{laina2016deeper}, while achieving comparable $\delta_1$, $\delta_2$ and $\delta_3$ accuracies.  Furthermore, it is important to note that these significant computational complexity improvements achieved by DepthNet Nano compared to~\cite{laina2016deeper} despite the fact that the input image size is $>$2$\times$ resolution both vertically and horizontally.

\textbf{Speed and energy efficiency}. The inference speed of DepthNet Nano and~\cite{alhashim2018high} are compared on a Jetson AGX Xavier embedded module in Table \ref{tab:comp-infer}.  For KITTI, DepthNet Nano was more than $4.6\times$ faster and more energy efficient than Alhashim et al.~\cite{alhashim2018high} at both 30W and 15W power budgets. For NYU Depth v2, DepthNet Nano are more than $2.3\times$ faster and more energy efficient than Alhashim et al.~\cite{alhashim2018high} at both 30W and 15W power budgets.  These quantitative results demonstrate that the proposed DepthNet Nano networks, created using a human-machine collaborative design strategy, can achieve a strong balance between accuracy, network architecture complexity, and computational complexity that makes it very well suited for embedded depth estimation for edge scenarios.

\begin{table}[]
\centering
\caption{\textbf{KITTI Inference.} All networks are evaluated on a Jetson AGX Xavier embedded module. Best results in \textbf{bold}.}
\begin{adjustbox}{max width=\linewidth}
\begin{tabular}{c||c|c|c|c|c|c}
\hline
Model & \begin{tabular}[c]{@{}c@{}}MACs\\ {[}G{]}\end{tabular} & $\delta_1$ & \begin{tabular}[c]{@{}c@{}}30W\\ {[}FPS{]}\end{tabular} & \begin{tabular}[c]{@{}c@{}}15W\\ {[}FPS{]}\end{tabular} &
\begin{tabular}[c]{@{}c@{}}30W\\ {[}$\frac{\text{images/s}}{\text{watt}}${]}\end{tabular} &
\begin{tabular}[c]{@{}c@{}}15W\\ {[}$\frac{\text{images/s}}{\text{watt}}${]}\end{tabular} \\ \hline \hline
Alhashim et al.~\cite{alhashim2018high} & 196 & 0.886 & 3.00 & 1.65 & 0.100 & 0.110 \\ \hline
DepthNet Nano & \textbf{4.66} & \textbf{0.894} & \textbf{13.92} & \textbf{7.70} & \textbf{0.464} & \textbf{0.513} \\ \hline
\end{tabular}%
\label{tab:comp-infer-kitti}
\end{adjustbox}
\end{table}

\begin{table}[]
\centering
\caption{\textbf{NYU Depth V2 Inference.} All networks are evaluated on a Jetson AGX Xavier embedded module. Best results in \textbf{bold}.}
\begin{adjustbox}{max width=\linewidth}
\begin{tabular}{c||c|c|c|c|c|c}
\hline
Model & \begin{tabular}[c]{@{}c@{}}MACs\\ {[}G{]}\end{tabular} & $\delta_1$ & \begin{tabular}[c]{@{}c@{}}30W\\ {[}FPS{]}\end{tabular} & \begin{tabular}[c]{@{}c@{}}15W\\ {[}FPS{]}\end{tabular} & \begin{tabular}[c]{@{}c@{}}30W\\ {[}$\frac{\text{images/s}}{\text{watt}}${]}\end{tabular} &
\begin{tabular}[c]{@{}c@{}}15W\\ {[}$\frac{\text{images/s}}{\text{watt}}${]}\end{tabular} \\ \hline \hline
Alhashim et al.~\cite{alhashim2018high} & 49.2 & \textbf{0.846} & 6.83 & 3.76 & 0.228 & 0.251 \\ \hline
DepthNet Nano & \textbf{4.4} & 0.816 & \textbf{15.8} & \textbf{8.8} & \textbf{0.527} & \textbf{0.587} \\ \hline
\end{tabular}%
\label{tab:comp-infer}
\end{adjustbox}
\end{table}

\subsection{Qualitative Analysis}
\begin{figure*}
  \includegraphics[width=\textwidth,height=2.5in]{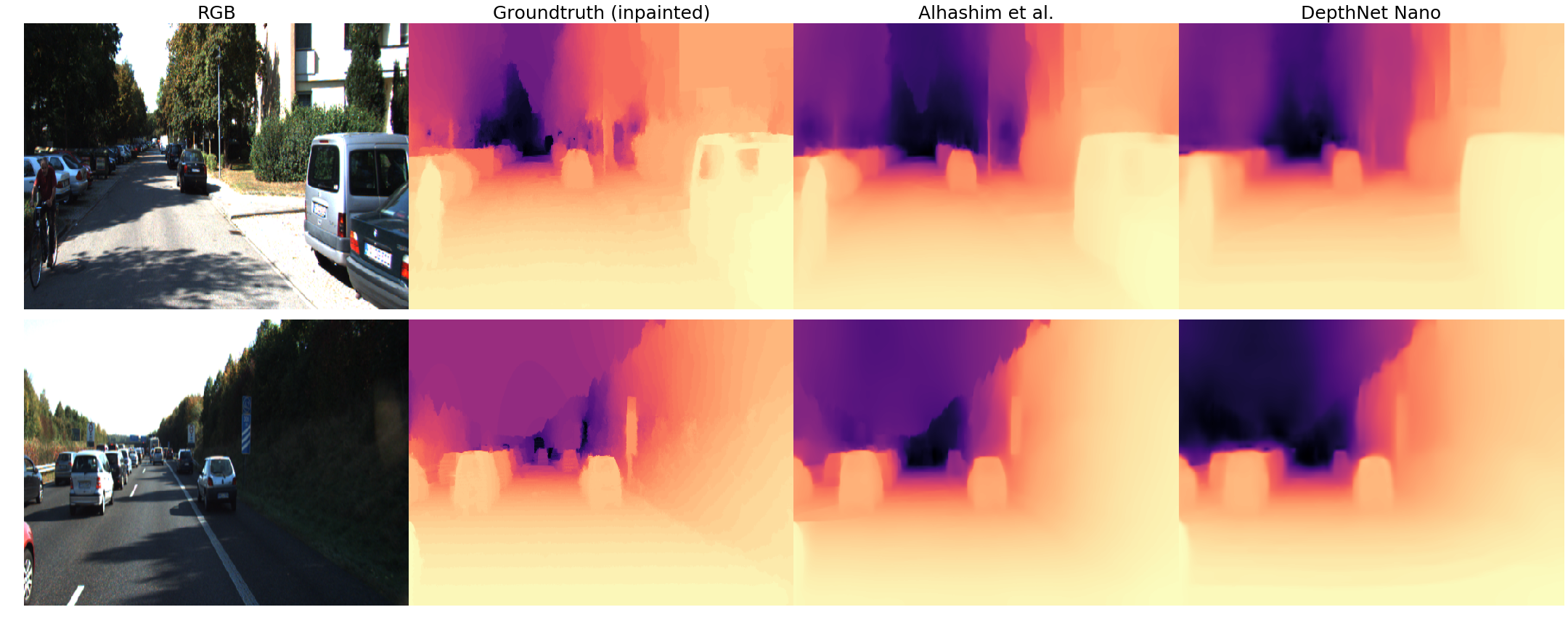}
  \caption{\textbf{Visualized results on KITTI dataset.} (Left to right) input RGB image, ground truth, and depth estimations from Alhashim et al.~\cite{alhashim2018high} and DepthNet Nano.  DepthNet Nano was able to produce high-quality depth estimations despite requiring 42$\times$ fewer MAC operations than Alhashim et al.~\cite{alhashim2018high}.}
  \label{fig:examples-kitti}
\end{figure*}

\begin{figure*}
  \includegraphics[width=\textwidth]{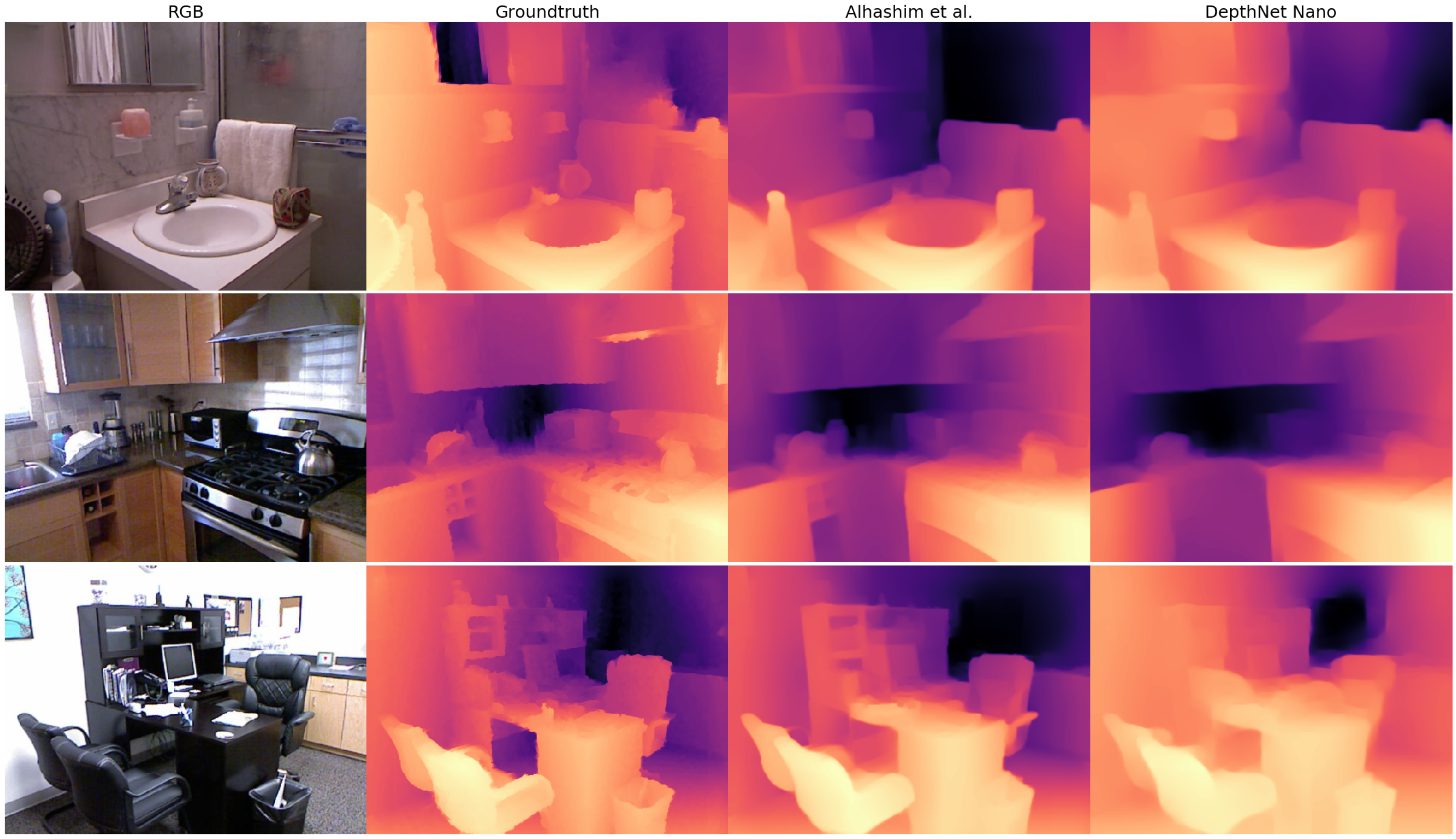}
  \caption{\textbf{Visualized results on NYU Depth V2 dataset.} (Left to right) input RGB image, ground truth, and depth estimations from Alhashim et al.~\cite{alhashim2018high} and DepthNet Nano.  DepthNet Nano was able to produce high-quality depth estimations despite requiring 10$\times$ fewer MAC operations than Alhashim et al.~\cite{alhashim2018high}.}
  \label{fig:examples}
\end{figure*}

In additional to quantitatively evaluating the performance DepthNet Nano, a qualitative analysis is also conducted to present areas that may be not be evaluated by the error metrics. Figures \ref{fig:examples-kitti} and \ref{fig:examples} shows three examples from the KITTI and NYU-Depth V2 datasets, respectively. Each RGB image has a corresponding ground-truth dense depth map and predicted dense depth maps from~\cite{alhashim2018high} and DepthNet Nano.  A number of interesting observations can be made based on the produced dense depth maps.  First of all, the dense depth maps produced by DepthNet Nano are visually comparable to that produced by~\cite{alhashim2018high} for KITTI and slightly less detailed than that produced by~\cite{alhashim2018high} for NYU Depth V2, despite DepthNet Nano requiring 42$\times$ and $\sim$10$\times$ fewer MAC operations, respectively.  Another interesting observation is the first row in Figure \ref{fig:examples}. In the ground-truth depth map, the shower curtain reflected in the mirror has a different depth than the mirror. Similarly, ~\cite{alhashim2018high} also predicted a different depth for the reflected shower curtain.  However, the proposed DepthNet Nano predicted a more consistent depth within the mirror. In this case, although the predicted depth maps for the mirror visually appears to be more accurate for DepthNet Nano, the result is not reflected when calculating error since the ground-truth depth map has an inconsistent depth for the mirror. Overall, visually, DepthNet Nano produced dense depth maps comparable to state of the art.

\section{Conclusion}
In this study, we introduced DepthNet Nano, a highly compact self normalizing network that is tailored for embedded monocular depth estimation and designed using a human machine collaborative design strategy. By coupling human-driven principled network design prototyping and machine-driven design exploration, the resulting DepthNet Nano network architecture exhibited highly customized macroarchitecture and microarchitecture designs, as well as self-normalizing characteristics that provide a strong balance between architecture complexity, computational complexity, and depth estimation performance.  Experimental results across two benchmark datasets demonstrated that the proposed DepthNet Nano possesses a significantly more architecturally and computationally efficient network architecture compared with state-of-the-art networks while achieving comparable performance on the NYU-Depth V2 and KITTI datasets. Furthermore, experiments demonstrated that DepthNet Nano had significantly faster inference speeds and energy efficiency on the Jetson AGX Xavier embedded module.  For future work, we plan to explore strategies for incorporating temporal information into the DepthNet Nano architecture in a way that improves performance while maintaining low architecture and computational complexity.   

{\small
\bibliographystyle{IEEEtran}
\bibliography{references}

\begin{thebibliography}{10}
\providecommand{\url}[1]{#1}
\csname url@samestyle\endcsname
\providecommand{\newblock}{\relax}
\providecommand{\bibinfo}[2]{#2}
\providecommand{\BIBentrySTDinterwordspacing}{\spaceskip=0pt\relax}
\providecommand{\BIBentryALTinterwordstretchfactor}{4}
\providecommand{\BIBentryALTinterwordspacing}{\spaceskip=\fontdimen2\font plus
\BIBentryALTinterwordstretchfactor\fontdimen3\font minus
  \fontdimen4\font\relax}
\providecommand{\BIBforeignlanguage}[2]{{%
\expandafter\ifx\csname l@#1\endcsname\relax
\typeout{** WARNING: IEEEtran.bst: No hyphenation pattern has been}%
\typeout{** loaded for the language `#1'. Using the pattern for}%
\typeout{** the default language instead.}%
\else
\language=\csname l@#1\endcsname
\fi
#2}}
\providecommand{\BIBdecl}{\relax}
\BIBdecl

\bibitem{fu2018deep}
H.~Fu, M.~Gong, C.~Wang, K.~Batmanghelich, and D.~Tao, ``Deep ordinal
  regression network for monocular depth estimation,'' in \emph{Proceedings of
  the IEEE Conference on Computer Vision and Pattern Recognition}, 2018, pp.
  2002--2011.

\bibitem{alhashim2018high}
I.~Alhashim and P.~Wonka, ``High quality monocular depth estimation via
  transfer learning,'' \emph{arXiv preprint arXiv:1812.11941}, 2018.

\bibitem{laina2016deeper}
I.~Laina, C.~Rupprecht, V.~Belagiannis, F.~Tombari, and N.~Navab, ``Deeper
  depth prediction with fully convolutional residual networks,'' in \emph{2016
  Fourth international conference on 3D vision (3DV)}.\hskip 1em plus 0.5em
  minus 0.4em\relax IEEE, 2016, pp. 239--248.

\bibitem{sandler2018mobilenetv2}
M.~Sandler, A.~Howard, M.~Zhu, A.~Zhmoginov, and L.-C. Chen, ``Mobilenetv2:
  Inverted residuals and linear bottlenecks,'' in \emph{Proceedings of the IEEE
  Conference on Computer Vision and Pattern Recognition}, 2018, pp. 4510--4520.

\bibitem{wofk2019fastdepth}
D.~Wofk, F.~Ma, T.-J. Yang, S.~Karaman, and V.~Sze, ``Fastdepth: Fast monocular
  depth estimation on embedded systems,'' \emph{arXiv preprint
  arXiv:1903.03273}, 2019.

\bibitem{yang2018netadapt}
T.-J. Yang, A.~Howard, B.~Chen, X.~Zhang, A.~Go, M.~Sandler, V.~Sze, and
  H.~Adam, ``Netadapt: Platform-aware neural network adaptation for mobile
  applications,'' in \emph{Proceedings of the European Conference on Computer
  Vision (ECCV)}, 2018, pp. 285--300.

\bibitem{wong2018ferminets}
A.~Wong, M.~J. Shafiee, B.~Chwyl, and F.~Li, ``Ferminets: Learning generative
  machines to generate efficient neural networks via generative synthesis,''
  \emph{arXiv preprint arXiv:1809.05989}, 2018.

\bibitem{wong2019yolo}
A.~Wong, M.~Famuori, M.~J. Shafiee, F.~Li, B.~Chwyl, and J.~Chung, ``Yolo nano:
  a highly compact you only look once convolutional neural network for object
  detection,'' \emph{arXiv preprint arXiv:1910.01271}, 2019.

\bibitem{he2016deep}
K.~He, X.~Zhang, S.~Ren, and J.~Sun, ``Deep residual learning for image
  recognition,'' in \emph{Proceedings of the IEEE conference on computer vision
  and pattern recognition}, 2016, pp. 770--778.

\bibitem{huang2017densely}
G.~Huang, Z.~Liu, L.~Van Der~Maaten, and K.~Q. Weinberger, ``Densely connected
  convolutional networks,'' in \emph{Proceedings of the IEEE conference on
  computer vision and pattern recognition}, 2017, pp. 4700--4708.

\bibitem{wong2019netscore}
A.~Wong, ``Netscore: Towards universal metrics for large-scale performance
  analysis of deep neural networks for practical on-device edge usage,'' in
  \emph{International Conference on Image Analysis and Recognition}.\hskip 1em
  plus 0.5em minus 0.4em\relax Springer, 2019, pp. 15--26.

\bibitem{klambauer2017self}
G.~Klambauer, T.~Unterthiner, A.~Mayr, and S.~Hochreiter, ``Self-normalizing
  neural networks,'' in \emph{Advances in neural information processing
  systems}, 2017, pp. 971--980.

\bibitem{mnas}
M.~Tan, B.~Chen, R.~Pang, V.~Vasudevan, and Q.~V. Le, ``Mnasnet: Platform-aware
  neural architecture search for mobile,'' \emph{arXiv preprint
  arXiv:1807.11626}, 2018.

\bibitem{fairnas}
X.~Chu, B.~Zhang, R.~Xu, and J.~Li, ``Fairnas: Rethinking evaluation fairness
  of weight sharing neural architecture search,'' \emph{arXiv preprint
  arXiv:1907.01845}, 2019.

\bibitem{eigen2014depth}
D.~Eigen, C.~Puhrsch, and R.~Fergus, ``Depth map prediction from a single image
  using a multi-scale deep network,'' in \emph{Advances in neural information
  processing systems}, 2014, pp. 2366--2374.

\bibitem{Adam}
D.~Kingma and J.~Ba, ``Adam: A method for stochastic optimization,''
  \emph{arXiv preprint arXiv:1412.6980}, 2017.

\bibitem{silberman2012indoor}
N.~Silberman, D.~Hoiem, P.~Kohli, and R.~Fergus, ``Indoor segmentation and
  support inference from rgbd images,'' in \emph{European Conference on
  Computer Vision}.\hskip 1em plus 0.5em minus 0.4em\relax Springer, 2012, pp.
  746--760.

\bibitem{geiger2012we}
A.~Geiger, P.~Lenz, and R.~Urtasun, ``Are we ready for autonomous driving? the
  kitti vision benchmark suite,'' in \emph{2012 IEEE Conference on Computer
  Vision and Pattern Recognition}.\hskip 1em plus 0.5em minus 0.4em\relax IEEE,
  2012, pp. 3354--3361.

\bibitem{levin2004colorization}
A.~Levin, D.~Lischinski, and Y.~Weiss, ``Colorization using optimization,'' in
  \emph{ACM SIGGRAPH 2004 Papers}, 2004, pp. 689--694.

\end{thebibliography}
}

\end{document}